\documentclass[sigconf]{acmart}

\acmConference[COLIEE 2026]{COLIEE 2026: 13th Competition on Legal Information Extraction and Entailment}{June 12, 2026}{Singapore}
\acmYear{2026}
\copyrightyear{2026}
\setcopyright{rightsretained}

\acmDOI{}
\acmISBN{}
\acmPrice{}

\usepackage{booktabs}
\usepackage{colortbl}
\usepackage{multirow}
\usepackage{graphicx}
\usepackage{amsmath}

\usepackage{amssymb}
\usepackage{xcolor}
\usepackage{hyperref}
\usepackage{CJKutf8}
\usepackage{tikz}
\usepackage{placeins}
\usetikzlibrary{positioning, arrows.meta, calc, fit, backgrounds, shadows}

\newcommand{\jp}[1]{\begin{CJK}{UTF8}{min}#1\end{CJK}}
\newcommand{\hlnum}[1]{\fcolorbox{red!70}{red!5}{\textbf{#1}}}

\definecolor{retrieval}{HTML}{2B6CB0}    
\definecolor{neural}{HTML}{C55A11}       
\definecolor{llm}{HTML}{9B2C2C}          
\definecolor{ensemble}{HTML}{553C9A}     
\definecolor{verdict}{HTML}{548235}      
\definecolor{lightbg}{HTML}{F5F5F5}      
\definecolor{shared}{HTML}{1B3A6B}
\definecolor{sharedlight}{HTML}{2B579A}
\definecolor{winbg}{HTML}{E8F5E9}  
\definecolor{postbg2}{HTML}{FFF3E0}  

\begin{document}

\title[Cross-Architecture LLM Ensembles, Feature-Based Reranking and Retrieval-Augmented Prompting]{Cross-Architecture LLM Ensembles, Feature-Based Reranking and Retrieval-Augmented Prompting for Legal Information Processing}

\author{Amal Saad Alshehri}
\affiliation{\institution{Durham University}\department{Department of Computer Science}\city{Durham}\country{United Kingdom}}
\affiliation{\institution{Jazan University}\department{Department of Computer Science}\city{Jazan}\country{Saudi Arabia}}
\email{ashahri@jazanu.edu.sa}

\author{Nelly Bencomo}
\affiliation{\institution{Durham University}\department{Department of Computer Science}\city{Durham}\country{United Kingdom}}
\email{nelly.bencomo@durham.ac.uk}

\author{Amir Atapour-Abarghouei}
\affiliation{\institution{Durham University}\department{Department of Computer Science}\city{Durham}\country{United Kingdom}}
\email{amir.atapour-abarghouei@durham.ac.uk}

\begin{abstract}
Legal information processing encompasses a heterogeneous set of retrieval, entailment and judgment prediction problems, requiring combinations of text matching, reasoning and robust generalisation with limited supervision. This paper presents a unified study of these challenges through a suite of open-weight systems spanning legal case retrieval, case entailment, statute retrieval and entailment and legal judgment prediction. In this paper, we report Team DU's participation in all five tasks of COLIEE 2026, with all systems relying exclusively on open-weight models. For Tasks 3 and 4, all models were released before 15 July 2025, as required by the competition rules.
For Task~4 (statute entailment), a cross-architecture ensemble of nine models from three families achieves 96.3\% accuracy, placing first among 33 submissions from 11 teams.
For the Pilot Task (tort prediction and rationale extraction), a multi-view system that combines five claim-level models and refines the case verdict using features derived from the claim predictions achieves 73.1\% TP accuracy and 68.2\% RE F1 as an unofficial submission, scoring above all official entries on TP and matching the highest on RE.
For Task~2 (legal case entailment), changing only the prompt instruction from single-selection to multi-selection raises F1 from 0.343 to 0.555 in post-competition evaluation on released gold labels, exceeding the best official submission (F1 = 0.490).
For Task~3 (statute retrieval and entailment), replacing the entailment model with Qwen3-235B and a structured legal reasoning prompt raises accuracy from 79.3\% to 91.5\% in post-competition analysis.
For Task~1 (legal case retrieval), a learning-to-rank system that combines lexical and semantic retrieval with structural, citation authority, and temporal features (34 in total) achieves F1 = 0.314 (rank 11 of 54 submissions from 22 teams). Taken together, these results show that legal information processing benefits from different forms of inductive bias across tasks, with cross-architecture ensembling, feature-based reranking and retrieval-augmented prompting each proving most effective in different settings.
\end{abstract}

\keywords{COLIEE, legal information retrieval, legal entailment, LLM ensemble, learning-to-rank, legal judgment prediction}

\maketitle

\section{Introduction}

Legal information processing involves a diverse family of problems, including document retrieval, textual entailment and judgment prediction, each of which places different demands on representation learning, structured reasoning and decision-making under domain-specific constraints \cite{alshehri2026neural}. These tasks become particularly challenging in legal settings, where documents are long, terminology is highly specialised, relevant evidence may be distributed unevenly across a text and small linguistic distinctions can materially alter legal meaning.

COLIEE\footnote{\url{https://coliee.org/}} provides a well-established benchmark for studying these challenges in a controlled setting and has run an annual competition on legal information extraction and entailment running since 2014~\cite{coliee2025overview,coliee2026overview}.
The 2026 edition~\cite{coliee2026overview} comprises four established tasks and one pilot task.
Tasks~1 and~2 address Canadian case law: Task~1 requires retrieving cases originally cited by a query case, and Task~2 requires identifying which paragraphs from a candidate case entail a given decision fragment.
Tasks~3 and~4 address Japanese statute law: Task~3 requires both retrieving relevant Civil Code articles and predicting entailment, while Task~4 tests entailment alone with the relevant articles provided by the organisers.
The Pilot Task addresses legal judgment prediction for Japanese tort cases, requiring both a binary verdict prediction (tort prediction, TP) and per-claim acceptance labels (rationale extraction, RE).
We participate in all five tasks as Team DU.
Table~\ref{tab:summary} summarises our submissions.
Our strongest results are on Task~4, where a cross-architecture ensemble achieves first place. On the Pilot Task, our unofficial submission achieves 73.1\% TP accuracy and 68.2\% RE F1, scoring above all official entries on TP and matching the highest on RE.
Post-competition analysis with the released gold labels identifies improvements for Tasks~2 and~3 that require no changes to the system architecture.

\begin{table}[t]
\centering
\caption{Summary of Team DU submissions across five tasks.}
\label{tab:summary}
\setlength{\tabcolsep}{2.5pt}
\footnotesize
\begin{tabular}{llcc}
\toprule
Task & Approach & Best Score & Rank \\
\midrule
1 & LightGBM LTR (34 features) & F1 = .314 & 11/54 \\
2 & MonoT5 + few-shot LLM & F1 = .343 & 22/35 \\
3 & BM25 bigrams + Qwen-72B & Acc = .793 & 14/22 \\
4 & Cross-arch meta-ensemble (DU1) & Acc = .963 & \textbf{1st} \\
Pilot & 5-view BERT + verdict bridge & TP = .731 & —$^*$ \\
\bottomrule
\multicolumn{4}{l}{\scriptsize $^*$Unofficial entry; TP = .731 exceeds all official entries and RE F1 = .682} \\
\multicolumn{4}{l}{\scriptsize \phantom{$^*$}matches the highest official result (see Section~\ref{sec:pilot}).} \\
\end{tabular} \vspace{-0.3cm}
\end{table}

The remainder of this paper is organised by task to reflect the heterogeneity of the benchmark.
Sections~\ref{sec:task1} through \ref{sec:pilot} describe the method, results, and analysis for each task.
Section~\ref{sec:discussion} discusses cross-task patterns and Section~\ref{sec:conclusion} concludes.

\section{Task 1: Legal Case Retrieval}
\label{sec:task1}

Legal case retrieval is challenging because cited authorities are often related by legal function rather than surface lexical overlap alone, while the underlying documents are long and structurally heterogeneous. Given a query case from the Canadian Federal Court, where all references to cited cases have been removed from the text, the objective of our system is to identify which cases in the corpus were originally cited by the query.
The organisers provide a training set of 2,001 labelled query cases over a corpus of 7,708 candidates.
The test set contains 400 queries over a separate corpus of 1,848 candidates.
Performance is measured by micro-averaged F1 at $k$=5, where each query receives exactly five predicted citations.

\subsection{Method}

\begin{figure}[t]
\centering
\resizebox{0.95\columnwidth}{!}{%
\begin{tikzpicture}[
    node distance=0.55cm,
    >={Stealth[length=4pt]},
    sbox/.style={draw=none, text=white, rounded corners=3pt,
                minimum height=0.7cm, align=center, font=\small\bfseries,
                minimum width=6.8cm},
    arr/.style={->, thick, color=black!50},
]
\node[sbox, fill=retrieval, minimum width=3.3cm] (bm25) at (0, 0) {BM25};
\node[sbox, fill=retrieval, minimum width=3.3cm, right=0.2cm of bm25] (dense) {Sentence-BERT};
\node[sbox, fill=retrieval!80, below=0.55cm of $(bm25.south)!0.5!(dense.south)$] (pool) {Candidate Pool ($\leq$1{,}000)};
\draw[arr] (bm25.south) -- (bm25.south |- pool.north);
\draw[arr] (dense.south) -- (dense.south |- pool.north);
\node[sbox, fill=neural, below=of pool] (feat) {34 features: 11 retrieval, 16 structural, 3 authority, 4 temporal};
\draw[arr] (pool) -- (feat);
\node[sbox, fill=llm, below=of feat] (lgb) {LightGBM LambdaRank};
\draw[arr] (feat) -- (lgb);
\node[sbox, fill=verdict, below=of lgb] (post) {Post-processing (self-match \& future-case removal)};
\draw[arr] (lgb) -- (post);
\end{tikzpicture}%
}
\caption{Task~1: Stage 1 retrieves candidates via BM25 and dense retrieval. Stage 2 extracts 34 features across four families and feeds them to a LightGBM reranker. Stage 3 enforces domain constraints.}
\label{fig:task1}
\end{figure}
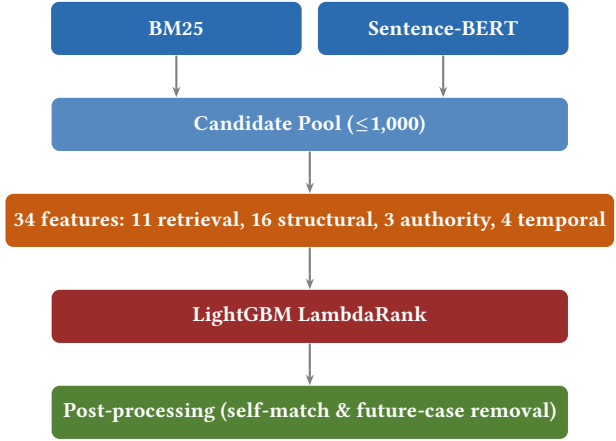

The system has three stages (Figure~\ref{fig:task1}).

\textbf{Stage 1: Hybrid candidate retrieval.}
BM25~\cite{robertson2009} and a Sentence-BERT dense retriever~\cite{reimers2019sbert} each return the top 1,000 candidates per query.
The union of both ranked lists forms a candidate pool containing 99.1\% of the true cited cases on the development set.
BM25 excels at matching specific legal terminology and case names, while dense retrieval captures semantic similarity when surface terms differ.

\textbf{Stage 2: Feature-based reranking.}
For each query-candidate pair, 34 features from four families are extracted.
The \emph{retrieval family} (11 features) includes normalised BM25 scores, dense cosine similarity, RRF~\cite{rrf} scores combining the BM25 and dense rankings, Jaccard similarity, overlap coefficient, and IDF-weighted overlap.
The \emph{structural family} (16 features) segments documents into Background, Issues, Analysis, and Decision sections using rule-based heading detection and positional heuristics, and computes paragraph-level Jaccard similarities (maximum, mean of the top three, and a coverage score) within and across sections.
The \emph{authority family} (3 features) captures normalised citation indegree from the training corpus, the difference in indegree between query and candidate, and log-transformed indegree, exploiting the observation that frequently cited cases are more likely to be cited again.
The \emph{temporal family} (4 features) encodes the signed and absolute year difference between filing dates, a binary indicator for the 1 to 15 year citation window, and a binary indicator for candidates decided after the query, reflecting that 82\% of real citations fall within this window.
These 34 features are passed to a LightGBM~\cite{lightgbm} model trained with a LambdaRank objective on the 2,001 queries from the COLIEE 2026 training set.

\textbf{Stage 3: Post-processing.}
Self-matches and candidates dated after the query are filtered out, contributing +2.7 pp to F1 on the development set.
Three runs are submitted: DU1 (1,000 trees, 63 leaves, learning rate 0.05), DU2 (score-level fusion of DU1 and DU3), and DU3 (2,000 trees, 127 leaves, learning rate 0.03).

\subsection{Results and Analysis}

Table~\ref{tab:task1} reports the official results.
DU3 achieves F1 = 0.3141, ranking 11th of 54 submissions from 22 teams.
The competition winner, NOWJ, achieves F1 = 0.4220, a gap of 10.8 pp.
DU2 (score-level fusion of the two tree configurations) achieves F1 = 0.3136, nearly identical to DU3, indicating that the two tree configurations make similar errors and fusion provides no additional diversity.
Post-competition scaling to 4,000 trees (DU9) raises F1 to 0.3456, reducing the gap to the winner by 3.15 pp.

The median document length of 4,573 tokens exceeds the 512-token limit of most encoder models.
We tested both a fine-tuned cross-encoder and a zero-shot large language model as alternatives to the feature-based reranker, but neither improved over LightGBM, because truncation discards the very sections (Analysis and Decision) that our structural features identify as most informative.

\textbf{Feature ablation.}
Table~\ref{tab:task1_ablation} reports a cumulative feature ablation on the development set.
Starting from BM25 scores alone (F1 = 16.50\%), adding retrieval features raises F1 to 27.56\% (+11.06 pp), structural features to 29.37\% (+1.81 pp), authority features to 30.49\% (+1.12 pp), and temporal features to 31.18\% (+0.69 pp).
Expanding to all available training data raises F1 to 35.70\% (+4.52 pp).
In the final model, the top five features by total gain are RRF at $k$=10 (14.1\%), RRF at $k$=5 (6.5\%), mean top-3 paragraph Jaccard (3.8\%), citation indegree (2.5\%), and maximum paragraph-level IDF overlap (2.0\%).
The dominance of the two RRF features confirms that combining lexical and semantic retrieval signals is the most informative input.

\begin{table}[t]
  \centering
  \caption{Task~1 official results (selected) and post-competition.}                                \label{tab:task1}
  \setlength{\tabcolsep}{4pt}                      
  \footnotesize                                      
  \begin{tabular}{rlccc}
  \toprule                                          
  Rank & Run & Prec. & Rec. & F1 \\                       
  \midrule
  1  & NOWJ   & .424 & .421 & .422 \\
  3  & JNLP   & .434 & .393 & .413 \\              
  \midrule
  \textbf{11} & \textbf{DU3} & \textbf{.295} & \textbf{.337} & \textbf{.314} \\                     
   & DU2 & .294 & .336 & .314 \\                         
   & DU1 & .285 & .325 & .304 \\
  \midrule                                          
  \multicolumn{5}{l}{\emph{Post-competition}} \\            
  & DU9 & .324 & .370 & \hlnum{.346} \\             
  \bottomrule                                               
  \end{tabular}                                        
  \end{table}

\textbf{Feature interaction effects.}
The two binary temporal indicators (whether the candidate falls within the 1--15 year citation window and whether the candidate postdates the query) have near-zero individual importance.
However, removing both together degrades F1 by 2.08 pp on the development set, a larger effect than the authority or temporal groups individually.
Binary features serve as split guides in gradient-boosted trees, allowing subsequent splits to learn separate thresholds for each partition.

\begin{table}[t]
\centering
\caption{Task~1 feature ablation on the development set. Features are added cumulatively.}
\label{tab:task1_ablation}
\setlength{\tabcolsep}{4pt}
\footnotesize
\begin{tabular}{lcc}
\toprule
Feature Set & F1 (\%) & $\Delta$ (pp) \\
\midrule
BM25 only & 16.50 & --- \\
+ Retrieval features (11) & 27.56 & +11.06 \\
+ Structural features (16) & 29.37 & +1.81 \\
+ Authority features (3) & 30.49 & +1.12 \\
+ Temporal features (4) & 31.18 & +0.69 \\
+ All available training data & 35.70 & +4.52 \\
\bottomrule
\end{tabular}
\end{table}

\section{Task 2: Legal Case Entailment}
\label{sec:task2}

Case entailment is particularly sensitive to granularity mismatch, since the system must align a short decision fragment with one or more entailing paragraphs embedded in a much larger candidate case. Given a decision fragment and candidate paragraphs from another case, the system must identify which paragraph(s) entail the decision.
The training set contains cases with gold paragraph labels.
The test set contains 100 cases with 294 gold paragraphs (average 2.94 per case), evaluated by F1.

\subsection{Method}

\begin{figure}[t]
\centering
\resizebox{0.95\columnwidth}{!}{%
\begin{tikzpicture}[
    node distance=0.4cm,
    >={Stealth[length=4pt]},
    stage/.style={draw=none, rounded corners=4pt, minimum height=0.7cm, text centered, fill=#1, text=white, font=\small\bfseries, minimum width=7cm},
    data/.style={draw=gray!50, rounded corners=4pt, minimum height=0.6cm, text centered, fill=white, font=\small, minimum width=7cm},
    detail/.style={font=\scriptsize, text=black!60, align=center},
    arr/.style={->, thick, color=black!40},
]
\node[data] (input) {\textbf{Test case (unsolved)} --- decision fragment $q$ + candidate paragraphs};
\node[stage=retrieval, below=of input] (bm25) {Stage 1: BM25 Retrieval};
\node[detail, below=0cm of bm25] (d1) {Retrieve top 100 paragraphs by lexical similarity to $q$};
\draw[arr] (input) -- (bm25);
\node[stage=neural, below=0.15cm of d1] (rerank) {Stage 2: Reranker Ensemble};
\node[detail, below=0cm of rerank] (d2) {Fine-tuned MonoT5 (weight 0.8) + Qwen3-Reranker (weight 0.2), keep top 20};
\draw[arr] (d1) -- (rerank);
\node[stage=verdict, below=0.15cm of d2] (fewshot) {Retrieval-Augmented Few-Shot Prompting};
\node[detail, below=0cm of fewshot] (d3) {BM25 retrieves 3 solved training cases with similar legal terminology to $q$.\\Each is formatted as (decision fragment, correct paragraph) and added to the prompt.};
\draw[arr] (d2) -- (fewshot);
\node[stage=llm, below=0.15cm of d3] (llm) {Stage 3: LLM Entailment Selection};
\node[detail, below=0cm of llm] (d4) {DeepSeek-V3, DeepSeek-R1, and LLaMA-3.3-70B each independently select paragraph(s)};
\draw[arr] (d3) -- (llm);
\node[stage=ensemble, below=0.15cm of d4] (vote) {Majority Vote / Intersection};
\node[detail, below=0cm of vote] (d5) {Keep paragraphs that $\geq$2 of 3 LLMs agree on};
\draw[arr] (d4) -- (vote);
\node[data, below=0.15cm of d5] (out) {\textbf{Selected entailing paragraph(s)}};
\draw[arr] (d5) -- (out);
\end{tikzpicture}%
}
\caption{Task~2: three-stage pipeline with retrieval-augmented few-shot prompting. Three solved training cases, retrieved by BM25 similarity, are provided to each LLM as demonstrations. No test labels are used.}
\label{fig:task2}
\end{figure}
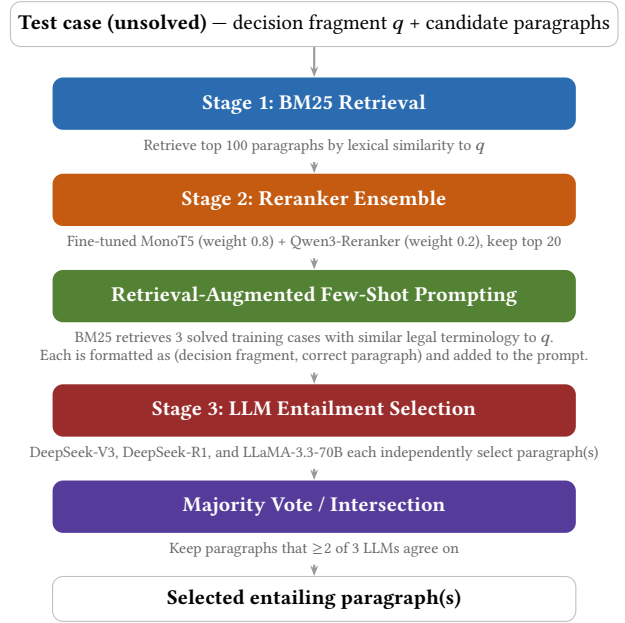

The system has three stages (Figure~\ref{fig:task2}).

\textbf{Stage 1: Lexical retrieval.}
BM25~\cite{robertson2009} retrieves the top 100 candidate paragraphs per decision fragment, achieving 99.0\% recall at rank 100 on the test set.

\textbf{Stage 2: Neural reranking.}
MonoT5-COLIEE~\cite{monot5}, a T5-based cross-encoder fine-tuned on the COLIEE Task~2 training data with hard negative mining, serves as the primary reranker.
Qwen3-Reranker-0.6B provides a complementary score, and the two are combined with weights 0.8 and 0.2 respectively.
The top 20 paragraphs are retained for the next stage.

\textbf{Stage 3: LLM entailment selection with retrieval-augmented few-shot prompting.}
A large language model receives the decision fragment, the reranked top-20 paragraphs, and three solved training examples retrieved by BM25 from an index built over all training queries.
By grounding the LLM in concrete precedent rather than relying on zero-shot reasoning, retrieval-augmented few-shot prompting adds 5.7 F1 points over zero-shot prompting on the development set.
Three models (DeepSeek-V3, DeepSeek-R1~\cite{deepseek-r1}, LLaMA-3.3-70B) independently select paragraphs.
Three runs use different aggregation strategies: DU1 (intersection of DeepSeek-V3 and R1), DU2 (DeepSeek-V3 alone, selected for its best precision-recall trade-off on the development set), DU3 (three-model tiebreaker without few-shot demonstrations, testing whether the models alone can compensate for the absence of retrieved examples).

\textbf{In-distribution hard negative training.}
A methodological contribution in Stage 2 is the MonoT5-v2 variant, which addresses a distribution mismatch in the standard hard negative mining approach.
MonoT5-v1 trains by pairing each correct paragraph with five incorrect paragraphs from the same case as negative examples.
At inference time, however, the model scores paragraphs that have already been reranked by the ensemble, which have a different score distribution from BM25 negatives.
MonoT5-v2 resolves this by using a two-pass approach: a first pass produces the reranker's own top-20 predictions on the training set, and the non-gold paragraphs from these predictions serve as hard negatives for a second pass of fine-tuning.
This ensures that the model trains on the exact distribution of candidates it will encounter at inference time.
MonoT5-v2 improves the reranker's ability to retain gold paragraphs in the top 20, reducing the number of gold paragraphs lost before they reach the LLM stage.

\subsection{Results and Analysis}

Table~\ref{tab:task2} reports the official results.
DU2 achieves the second-highest precision among all 35 submissions (0.753) but low recall (0.218), resulting in F1 = 0.338.
The competition winner, IAI, achieves F1 = 0.490.
The single-selection prompt was calibrated on the development set, where 80\% of cases have exactly one correct paragraph (average 1.22).
The test set averages 2.94 correct paragraphs per case, with 95\% having more than one, capping the theoretical single-selection F1 ceiling at 0.508.

\textbf{Development versus test distribution shift.}
DU2 achieves F1 = 0.764 on the development set but only F1 = 0.338 on the test set, a 42.6 pp gap entirely attributable to the distribution shift in the number of gold paragraphs per case.
This experience motivates a general recommendation: when the number of correct answers per query is uncertain, the safest strategy is to allow multi-selection with a confidence threshold rather than enforcing a fixed count.

\textbf{Post-competition multi-selection.}
Table~\ref{tab:task2_post} reports the effect of changing only the prompt instruction to permit multi-selection, with every other component identical.
Under majority voting across three models, this achieves F1 = 0.555 in post-competition evaluation on released gold labels, exceeding the best official submission by 6.5 points.
The entire gap is attributable to a single prompt instruction.

\textbf{Recall loss decomposition.}
Of the 230 missed gold paragraphs in the submitted single-selection system, 122 (53\%) are missed because the single-selection constraint prevents returning more than one paragraph per case even when the system has ranked multiple correct paragraphs highly.
A further 61 (27\%) are missed because the system selects an incorrect paragraph, and 47 (20\%) are missed because the LLM returns no selection at all.
The single-selection constraint accounts for the majority of the recall loss.

\begin{table}[t]
\centering
\caption{Task~2 results: official submissions and post-competition multi-selection (same pipeline, different prompt instruction).}
\label{tab:task2}
\label{tab:task2_post}
\setlength{\tabcolsep}{3pt}
\footnotesize
\begin{tabular}{lccc}
\toprule
Configuration & Prec. & Rec. & F1 \\
\midrule
\multicolumn{4}{l}{\emph{Official submissions}} \\
IAI (winner)                          & .450 & .537 & .490 \\
\textbf{DU3}                          & .691 & .228 & .343 \\
\textbf{DU2}                          & \textbf{.753} & .218 & .338 \\
\textbf{DU1}                          & .663 & .214 & .324 \\
\midrule
\multicolumn{4}{l}{\emph{Post-competition (multi-selection prompt)}} \\
DU majority vote                      & .619 & .503 & \hlnum{.555} \\
\bottomrule
\end{tabular}
\end{table}

Under multi-selection, DeepSeek-V3 with few-shot prompting achieves F1 = 0.549, DeepSeek-R1 achieves F1 = 0.501, and DeepSeek-V3 without few-shot achieves F1 = 0.533, while the majority vote of all three (F1 = 0.555) exceeds every individual model, confirming complementary selection errors across the three models.

\section{Task 3: Statute Retrieval and Entailment}
\label{sec:task3}

Statute retrieval and entailment combines two sources of difficulty: identifying the correct provisions from a compact but highly cross-referential code base, and then reasoning over those provisions with precise control of legal language. Task~3 requires retrieving relevant articles from the 725-article Japanese Civil Code and predicting entailment (Y/N) for 82 test queries from the bar examination.
The training set comprises 965 queries from the Japanese bar examination, each annotated with the relevant article(s) and a Y/N entailment label.

\subsection{Method}

\begin{figure}[t]
\centering
\resizebox{0.98\columnwidth}{!}{%
\begin{tikzpicture}[
    node distance=0.35cm,
    >={Stealth[length=3pt]},
    sbox/.style={draw=none, fill=retrieval, text=white, rounded corners=3pt,
                minimum width=7.2cm, align=center, inner sep=4pt, font=\small\bfseries},
    exbox/.style={draw=retrieval!40, fill=white, rounded corners=2pt,
                minimum width=6.8cm, align=left, inner sep=3pt,
                font=\scriptsize, text width=6.5cm},
    runbox/.style={rounded corners=3pt,
                minimum height=0.6cm, minimum width=7.2cm, align=left,
                font=\small, text width=6.8cm, inner sep=3pt},
    arr/.style={->, thick, color=black!40},
  ]
  \node[sbox] (q) {82 Test Queries};
  \node[sbox, below=0.4cm of q] (bm) {BM25 Character-Bigram Retrieval (top 10)};
  \node[exbox, below=0.1cm of bm] (bmex) {%
    \texttt{Query: \jp{検察官は後見開始の審判を請求する}}\\[1pt]
    \texttt{Bigrams: \jp{検察} $|$ \jp{察官} $|$ \jp{後見} $|$ \jp{審判} $|$ \jp{請求} $|$ \ldots}\\[1pt]
    \texttt{Art.~7 matches: \jp{後見}\,(guard.) \jp{審判}\,(adjud.) \jp{請求}\,(petition)}};
  \node[sbox, below=0.3cm of bmex] (xr) {Cross-Reference Expansion (+1--7 articles)};
  \node[exbox, below=0.1cm of xr] (xrex) {%
    \texttt{Art.~28: \jp{``第百三条に規定する権限を超える...''}}\\[1pt]
    \texttt{$\to$ Detects citation, adds Art.~103 to candidates}};
  \draw[arr] (q)--(bm); \draw[arr] (bmex)--(xr);
  \node[font=\scriptsize\itshape, text=black!40, below=0.3cm of xrex] (lb) {Entailment};
  \node[runbox, draw=retrieval!50, fill=retrieval!10, below=0.1cm of lb] (du) {%
    \textbf{DU}\enspace Top 5 articles $\to$ Qwen2.5-72B, single call ($T{=}0$) $\to$ Y/N};
  \node[runbox, draw=neural!50, fill=neural!10, below=0.15cm of du] (du2) {%
    \textbf{DU2}\enspace 2\,400-char budget $\to$ Qwen2.5-72B, structured JSON ($T{=}0$) $\to$ Y/N};
  \node[runbox, draw=verdict!50, fill=verdict!10, below=0.15cm of du2] (du1) {%
    \textbf{DU1}\enspace 3-way RRF $\to$ Qwen-7B rerank $\to$ Qwen-72B SC vote ($k{=}5$) $\to$ Y/N};
  \draw[arr] (xrex.south) -- (lb.north);
  \begin{scope}[on background layer]
    \node[draw=retrieval!30, fill=retrieval!4, rounded corners=5pt,
          fit=(q)(bm)(bmex)(xr)(xrex), inner sep=5pt] {};
  \end{scope}
\end{tikzpicture}%
}
\caption{Task~3: BM25 matches queries and articles using character bigrams. Cross-reference expansion detects citations between articles. Three entailment configurations use Qwen2.5-72B with different prompting strategies.}
\label{fig:task3}
\end{figure}
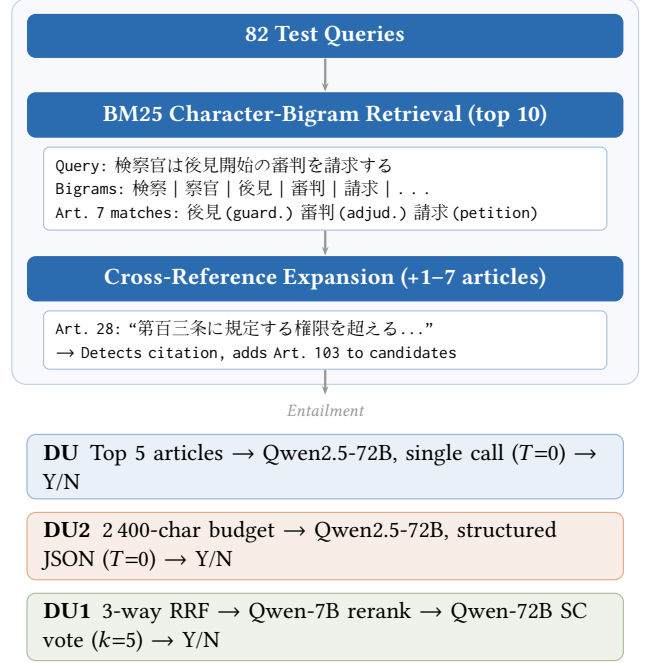

The system has two stages (Figure~\ref{fig:task3}).

\textbf{Retrieval: BM25 with character bigrams.}
Because Japanese text does not use spaces between words, standard BM25 cannot split text into tokens directly. Instead of relying on a morphological analyser, which can mis-segment specialised legal terms, we split every text into overlapping pairs of consecutive characters (character bigrams). For example, \begin{CJK}{UTF8}{min}後見開始\end{CJK} (commencement of guardianship) produces the bigrams \begin{CJK}{UTF8}{min}後見\end{CJK}, \begin{CJK}{UTF8}{min}見開\end{CJK}, \begin{CJK}{UTF8}{min}開始\end{CJK}. BM25~\cite{robertson2009} then matches queries and articles by shared bigrams and retrieves the top 10 candidates.

\textbf{Cross-reference expansion.}
Civil Code articles frequently cite other articles by number.
If BM25 retrieves an article that references another, we detect the citation using regular expressions and add the referenced article to the candidate pool, typically expanding from 10 to 11--17 articles per query.
Cross-reference expansion captures legally relevant articles that lack surface-text overlap with the query.

\textbf{Entailment.}
Three runs use Qwen2.5-72B-Instruct~\cite{qwen2.5}: DU reads the top 5 articles in a single prompt, DU2 limits context to 2,400 characters with structured JSON output (achieving the best accuracy at 79.3\%), and DU1 fuses BM25, TF-IDF, and BGE-M3 dense embeddings~\cite{bge-m3} via RRF~\cite{rrf} with self-consistency voting~\cite{wang2023self}.
Despite its additional complexity, DU1 achieves the lowest accuracy (75.6\%), illustrating that pipeline complexity can degrade performance when the additional components introduce noise.

\subsection{Results and Analysis}

Table~\ref{tab:task3} reports the official results.
DU2 achieves 65/82 correct (79.3\%, rank 14 of 22 submissions).
The competition winner, NOWJ, achieves 78/82 (95.1\%).
Retrieval recall ranges from 0.94 to 0.98 across the three runs, confirming that the entailment model rather than retrieval is the primary bottleneck.
Error analysis reveals a systematic positive bias: of the 14 errors DU2 makes, 12 are false positives (wrong Y) and only 2 are false negatives (wrong N).

\textbf{Ensemble prediction balance.}
The 9-expert ensemble applied post-competition predicts 42Y/40N on the 82 test queries, closely tracking the gold distribution of 41Y/41N and reducing the false-positive count from 12 to 6 compared to DU2's 51Y/31N prediction.

\textbf{Post-competition ensemble.}
Table~\ref{tab:task3_post} reports results from applying our Task~4 ensemble (Section~\ref{sec:task4}), the top-scoring system on the Task~4 leaderboard, to the same retrieved articles.
Qwen3-235B with IRAC prompting achieves 91.5\% (75/82 correct), while the nine-expert majority vote achieves 86.6\% (71/82).
The improvement from 79.3\% to 91.5\% represents a gain of +12.2 pp from replacing the model alone.
Notably, the 9-expert ensemble (86.6\%) underperforms the single Qwen3-235B model (91.5\%) on Task~3, unlike Task~4, where the ensemble dominates.
This reversal occurs because the Task~3 candidate pool includes distractor articles that confuse weaker ensemble members, and their incorrect votes dilute the signal from the stronger experts.

\begin{table}[t]
\centering
\caption{Task~3 results: official submissions and post-competition analysis on the same retrieved articles.}
\label{tab:task3}
\label{tab:task3_post}
\setlength{\tabcolsep}{4pt}
\footnotesize
\begin{tabular}{lccr}
\toprule
Configuration & Correct & Acc.\,(\%) & Rank \\
\midrule
\multicolumn{4}{l}{\emph{Official submissions (rank/22)}} \\
NOWJ\_run1                   & 78/82 & 95.1 & 1 \\
\textbf{DU2}                 & 65/82 & \textbf{79.3} & \textbf{14} \\
\textbf{DU}                  & 63/82 & \textbf{76.8} & \textbf{18} \\
\textbf{DU1}                 & 62/82 & \textbf{75.6} & \textbf{19} \\
\midrule
\multicolumn{4}{l}{\emph{Post-competition (same retrieved articles)}} \\
Qwen3-235B (IRAC)            & 75/82 & \hlnum{91.5} & --- \\
9-expert ensemble (DU3)      & 71/82 & 86.6 & --- \\
\bottomrule
\end{tabular}
\end{table}

\section{Task 4: Statute Entailment}
\label{sec:task4}

By removing the retrieval stage, Task 4 isolates the pure entailment problem and provides a controlled setting for analysing how different model families reason over statutory text. This task provides the relevant Civil Code articles and requires only the entailment decision (Y/N) for 82 test queries from the bar examination.
Unlike Task~3, where the system must first retrieve the relevant articles, Task~4 isolates the entailment component by providing gold articles as input.
The gold label distribution is approximately balanced, consistent across all exam years.
This section presents our first-place system.

\subsection{Expert Model Selection}

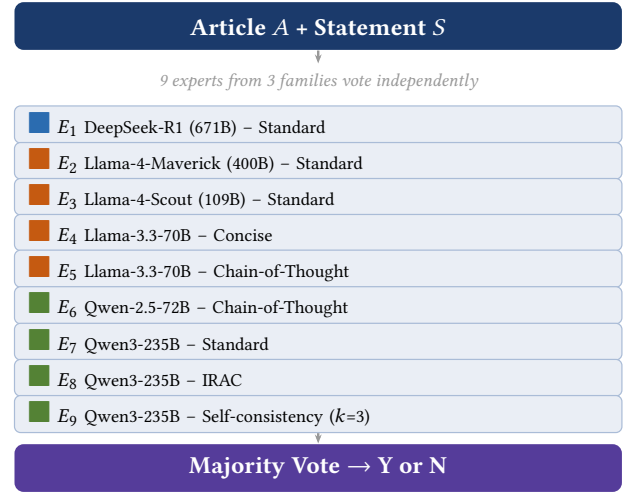
\begin{figure}[t]
\centering
\resizebox{0.95\columnwidth}{!}{%
\begin{tikzpicture}[
    node distance=0.25cm,
    >={Stealth[length=3pt]},
    sbox/.style={draw=none, text=white, rounded corners=3pt,
                minimum height=0.55cm, minimum width=7cm, align=center, font=\small\bfseries},
    ebox/.style={draw=sharedlight!40, fill=sharedlight!10, rounded corners=2pt,
                minimum height=0.38cm, minimum width=7cm, align=left,
                font=\scriptsize, inner sep=2pt, text width=6.7cm},
    arr/.style={->, thick, color=black!40},
  ]
  \node[sbox, fill=shared] (input) {Article $A$ + Statement $S$};
  \node[font=\scriptsize\itshape, text=black!40, below=0.15cm of input] (lbl) {9 experts from 3 families vote independently};
  \node[ebox, below=0.06cm of lbl] (e1) {{\color{retrieval}\large$\blacksquare$} $E_1$ DeepSeek-R1 (671B) -- Standard};
  \node[ebox, below=-0.02cm of e1] (e2) {{\color{neural}\large$\blacksquare$} $E_2$ Llama-4-Maverick (400B) -- Standard};
  \node[ebox, below=-0.02cm of e2] (e3) {{\color{neural}\large$\blacksquare$} $E_3$ Llama-4-Scout (109B) -- Standard};
  \node[ebox, below=-0.02cm of e3] (e4) {{\color{neural}\large$\blacksquare$} $E_4$ Llama-3.3-70B -- Concise};
  \node[ebox, below=-0.02cm of e4] (e5) {{\color{neural}\large$\blacksquare$} $E_5$ Llama-3.3-70B -- Chain-of-Thought};
  \node[ebox, below=-0.02cm of e5] (e6) {{\color{verdict}\large$\blacksquare$} $E_6$ Qwen-2.5-72B -- Chain-of-Thought};
  \node[ebox, below=-0.02cm of e6] (e7) {{\color{verdict}\large$\blacksquare$} $E_7$ Qwen3-235B -- Standard};
  \node[ebox, below=-0.02cm of e7] (e8) {{\color{verdict}\large$\blacksquare$} $E_8$ Qwen3-235B -- IRAC};
  \node[ebox, below=-0.02cm of e8] (e9) {{\color{verdict}\large$\blacksquare$} $E_9$ Qwen3-235B -- Self-consistency ($k$=3)};
  \node[sbox, fill=ensemble, below=0.12cm of e9] (vote) {Majority Vote $\to$ Y or N};
  \draw[arr] (input) -- (lbl);
  \draw[arr] (e9.south) -- (vote.north);
\end{tikzpicture}%
}
\caption{Task~4 base 9-expert ensemble (submitted as DU3; reused inside DU1's meta-ensemble). Three families --- {\color{retrieval}\large$\blacksquare$}~DeepSeek, {\color{neural}\large$\blacksquare$}~Llama, {\color{verdict}\large$\blacksquare$}~Qwen --- vote independently on Y/N.}
\label{fig:task4}
\end{figure}

We assemble nine expert configurations from three model families: DeepSeek~\cite{deepseek-r1}, Llama~\cite{llama3}, and Qwen~\cite{qwen2.5}, spanning two architecture types (Mixture-of-Experts and Dense Transformer).
The selection criteria are: architectural diversity across families, complementary prompting strategies, and individually competitive zero-shot performance.
The three model families collectively span parameter counts from 70B (Llama-3.3-70B Dense) to 671B (DeepSeek-R1 MoE), providing diversity not only in training data and architecture but also in model scale.
Table~\ref{tab:experts} lists all nine configurations.
Individual accuracies on the 258-question validation benchmark (H30+R01+R02) range from 80.2\% to 84.9\%, while the unweighted 9-expert majority vote (DU3) reaches 91.9\% and the DU1 meta-ensemble built on top of it reaches 93.0\%.
Figure~\ref{fig:task4} illustrates the architecture.

\begin{table}[t]
\caption{The 9 experts submitted as DU3 (also reused inside DU1). Validation accuracy (\%); per-split in Table~\ref{tab:task4_val}.}
\label{tab:experts}
\setlength{\tabcolsep}{2pt}
\footnotesize
\begin{tabular}{llllr}
\toprule
 & Model & Arch. & Prompt & Val. \\
\midrule
$E_1$ & DeepSeek-R1 (671B) & MoE & Standard & 84.1 \\
$E_2$ & Llama-4-Maverick (400B) & MoE & Standard & 83.7 \\
$E_3$ & Llama-4-Scout (109B) & MoE & Standard & 83.3 \\
$E_4$ & Llama-3.3-70B & Dense & Concise & 80.2 \\
$E_5$ & Llama-3.3-70B & Dense & CoT Strict & 80.6 \\
$E_6$ & Qwen-2.5-72B & Dense & CoT Strict & 82.9 \\
$E_7$ & Qwen3-235B & MoE & Standard & 84.1 \\
$E_8$ & Qwen3-235B & MoE & IRAC & 84.5 \\
$E_9$ & Qwen3-235B & MoE & SC ($k$=3) & 84.9 \\
\midrule
\multicolumn{4}{l}{\textbf{DU3 (9-expert majority vote)}} & \textbf{91.9} \\
\multicolumn{4}{l}{\textbf{DU1 (meta-ensemble, see Section~\ref{sec:task4})}} & \textbf{93.0} \\
\bottomrule
\end{tabular}
\end{table}

\subsection{Expert Instruction Design}

Each expert receives a Japanese-language system prompt specifying the reasoning procedure.
Table~\ref{tab:prompts} lists the five prompt variants in both Japanese and English.
The core \textit{Standard} prompt instructs the model to: (1) identify the article's requirements precisely, (2) verify exception clauses, (3) check negation consistency, (4) examine quantifier expressions, and (5) determine entailment based exclusively on the provided text.
The \textit{Concise} prompt removes the multi-step instruction and asks for only the final label.
The \textit{CoT Strict} prompt enforces a rigid six-step chain-of-thought procedure.
The \textit{IRAC} prompt follows the standard legal reasoning framework (Issue, Rule, Application, Conclusion).
The \textit{Self-consistency} variant generates $k$=3 independent responses at temperature 0.7 and returns the majority answer~\cite{wang2023self}.
All prompts include an explicit instruction that the label distribution is approximately 50\% Y and 50\% N, and that the model should not be biased towards either label. This debiasing statement reduces positive bias on the validation set.
The choice of Japanese as the prompt language is deliberate: language-matched prompts avoid translation-induced errors in legal terminology.

\begin{table}[t]
\caption{Prompt variants for Task~4. All prompts include an explicit debiasing statement. Japanese originals shown with English translations.}
\label{tab:prompts}
\setlength{\tabcolsep}{3pt}
\scriptsize
\begin{tabular}{p{1.2cm}p{5.8cm}}
\toprule
Variant & Instruction \\
\midrule
Standard &
\jp{判断の手順：1.条文の要件・条件を正確に読み取る 2.例外規定・ただし書きを確認する 3.否定の整合性を確認する 4.量化表現・限定語を確認する 5.条文の論理的帰結として導かれる場合はYと判定する} \\
& \textit{(1) Read requirements precisely. (2) Check exceptions. (3) Verify negation consistency. (4) Check quantifiers. (5) Judge Y if the statement follows logically.} \\
\midrule
Concise &
\jp{条文のみを根拠に、陳述文が論理的に正しいかを判定してください。正しければY、正しくなければNを最後の行に出力してください。} \\
& \textit{Based solely on the article, determine if the statement is logically correct. Output Y or N.} \\
\midrule
CoT Strict &
\jp{ステップ1：条文の要件を箇条書きで列挙 ステップ2：陳述文の主張を箇条書きで列挙 ステップ3：各要件との整合性を確認 ステップ4：ただし書き・例外を確認 ステップ5：限定語に注意 ステップ6：総合判定} \\
& \textit{Step 1: List requirements. Step 2: List claims. Step 3: Check consistency. Step 4: Check exceptions. Step 5: Note quantifiers. Step 6: Judgment.} \\
\midrule
IRAC &
\jp{IRAC法を用いて分析：1.Issue（争点）2.Rule（法規範）3.Application（適用）4.Conclusion（結論）} \\
& \textit{(1) Issue. (2) Rule. (3) Application. (4) Conclusion.} \\
\midrule
SC & Standard prompt, $k$=3 at $T$=0.7, majority vote. \\
\bottomrule
\end{tabular}
\end{table}

\subsection{Ensemble Construction and Submitted Runs}

Three runs were submitted, layering progressively more aggregation logic on top of a shared pool of (model, prompt) experts.

\textbf{DU3} is an unweighted 9-expert majority vote, with the expert subset (Table~\ref{tab:runs}, ``DU3'' column) selected by exhaustive combinatorial search over the validation benchmark. DU3 achieves 91.9\% on validation (Table~\ref{tab:task4_val}) and 77/82 = 93.9\% on the test set.

\textbf{DU2} also uses a 9-vote majority, but two of the nine votes are deliberation-derived. When a base ensemble's vote margin is at most one, three judge models (Qwen3-235B, DeepSeek-R1 and Llama-4-Maverick) re-evaluate the question with detailed reasoning prompts; their unanimous consensus and majority consensus form two additional ``deliberation experts'' that participate in the final vote alongside seven direct experts. DU2 achieves 92.6\% on validation and 79/82 = 96.3\% on the test set.

\textbf{DU1} (primary) is a hierarchical meta-ensemble: three sub-ensembles vote independently and the final DU1 prediction is the majority across their outputs. The sub-ensembles are Sub-A (8 experts plus a temporal-aware Qwen3-235B expert on R01 only; Table~\ref{tab:runs}, ``Sub-A'' column), Sub-B (5 experts; ``Sub-B'' column), and Sub-C, which is architecturally identical to DU2. DU1 achieves 93.0\% on validation and 79/82 = 96.3\% on the test set --- the highest accuracy among 33 submissions from 11 teams.

The best single expert ($E_9$, Qwen3-235B with self-consistency) achieves 84.9\% overall, a gap of 8.1 pp below DU1, demonstrating the substantial contribution of cross-architecture voting and hierarchical aggregation.
Plain 9-expert majority voting (DU3, 91.9\%) outperformed all five learned combination methods tested (logistic regression, random forest, gradient boosting, SVM, MLP), which ranged from 85.3\% to 90.3\%. With only 258 validation questions, the training folds provide insufficient data for reliable parameter estimation.

\begin{table}[t]
\caption{Validation accuracy per split for each submitted run and the best single expert.}
\label{tab:task4_val}
\setlength{\tabcolsep}{4pt}
\footnotesize
\begin{tabular}{lcccc}
\toprule
Run & H30 & R01 & R02 & Overall \\
\midrule
DU3 (plain 9-vote)         & 94.0 & 87.3 & 96.3 & 91.9 \\
DU2 (9-vote w/ deliberation) & 94.0 & 88.2 & 97.5 & 92.6 \\
DU1 (meta-ensemble)        & 94.0 & 88.2 & 98.8 & 93.0 \\
Best single ($E_9$) & 91.0 & 78.2 & 88.9 & 84.9 \\
\bottomrule
\end{tabular}
\end{table}

\begin{table}[t]
\caption{Expert membership: DU3 (plain 9-vote) and two of DU1's three sub-ensembles (Sub-A, Sub-B). DU1's third sub-ensemble is identical to DU2 (Section~\ref{sec:task4}). Run accuracies in Table~\ref{tab:task4_val}.}
\label{tab:runs}
\small
\setlength{\tabcolsep}{2.5pt}
\begin{tabular}{llccc}
\toprule
Model Family & Prompt & DU3 & Sub-A & Sub-B \\
\midrule
DeepSeek-R1 & Standard & $\checkmark$ & & \\
DeepSeek-R1 & Meticulous & & & $\checkmark$ \\
Llama-4-Maverick & Standard & $\checkmark$ & $\checkmark$ & \\
Llama-4-Scout & Standard & $\checkmark$ & $\checkmark$ & \\
Llama-3.3-70B & Standard & & & $\checkmark$ \\
Llama-3.3-70B & Concise & $\checkmark$ & & \\
Llama-3.3-70B & CoT & $\checkmark$ & $\checkmark$ & \\
Llama-3.3-70B & English & & $\checkmark$ & \\
Qwen-2.5-72B & CoT & $\checkmark$ & $\checkmark$ & $\checkmark$ \\
Qwen-2.5-72B & IRAC & & & $\checkmark$ \\
Qwen3-235B & Standard & $\checkmark$ & $\checkmark$ & $\checkmark$ \\
Qwen3-235B & IRAC & $\checkmark$ & $\checkmark$ & \\
Qwen3-235B & SC ($k$=3) & $\checkmark$ & $\checkmark$ & \\
\midrule
\multicolumn{2}{l}{Total experts} & 9 & 8$^{*}$ & 5 \\
\bottomrule
\multicolumn{5}{l}{\scriptsize $^{*}$Sub-A adds a temporal-aware Qwen3-235B expert on R01 only.}
\end{tabular}
\end{table}

\subsection{Results}

Table~\ref{tab:task4} reports the official test results.
DU1 and DU2 both achieve 79/82 (96.3\%), the highest accuracy among 33 submissions from 11 teams.
The best competitor achieves 78/82 (95.1\%).
The three errors involve complex interactions between exception clauses and quantifier expressions.
Two are false negatives where a majority of experts misinterpret a conditional exception beginning with \jp{ただし} (``provided that'') as a negation of the main rule.
The third error is a false positive caused by a subtle quantifier distinction between \jp{すべての} (``all'') and \jp{いずれかの} (``any'') that none of the nine experts detects, resulting in a unanimous but incorrect Y judgment.

\begin{table}[t]
\centering
\caption{Task~4 official test results (selected).}
\label{tab:task4}
\setlength{\tabcolsep}{4pt}
\footnotesize
\begin{tabular}{rlc}
\toprule
Rank & Run & Accuracy \\
\midrule
\textbf{1} & \textbf{DU1} & \hlnum{96.3\%} \\
\textbf{1} & \textbf{DU2} & \hlnum{96.3\%} \\
3 & JNLP-3 & 95.1\% \\
3 & IAIrun2 & 95.1\% \\
6 & \textbf{DU3} & \textbf{93.9\%} \\
\bottomrule
\end{tabular}
\end{table}

\subsection{Cross-Architecture versus Same-Model Ensembles}

Table~\ref{tab:gap} compares same-model and cross-architecture ensembles. The best same-model ensemble (Llama-3.3-70B with three prompts) reaches 87.2\% on the validation set.
A best-of-search 5-expert cross-architecture ensemble reaches 91.9\%, the 9-expert plain vote (DU3) also reaches 91.9\%, and the DU1 meta-ensemble built on top of the 9-vote layer reaches 93.0\%, demonstrating a 4.7 pp gap attributable to cross-architecture diversity over same-model voting and a further 1.1 pp from hierarchical aggregation.
Cross-family expert pairs disagree on 17.5\% of questions versus 11.4\% for same-family pairs, confirming that models from different families make more independent errors, which is the fundamental condition for effective majority voting.

\begin{table}[t]
\centering
\caption{Same-model versus cross-architecture ensemble accuracy on the 258-question validation benchmark.}
\label{tab:gap}
\setlength{\tabcolsep}{3pt}
\footnotesize
\begin{tabular}{lr}
\toprule
Configuration & Accuracy \\
\midrule
Best single expert & 84.9\% \\
Best same-model ensemble (Llama-3.3, 3 prompts) & 87.2\% \\
Best 5-expert cross-arch ensemble (search-selected) & 91.9\% \\
9-expert cross-arch plain vote (DU3) & 91.9\% \\
DU1 meta-ensemble (hierarchical) & 93.0\% \\
\bottomrule
\end{tabular}\vspace{-0.5cm}
\end{table}

\subsection{Analysis}

\textbf{Expert removal.}
The most impactful single removal from DU1 is Llama-4-Maverick ($-3.9$ pp), followed by Qwen-2.5-72B ($-3.5$ pp).
Removing all three Qwen3-235B variants reduces accuracy by 3.9 pp (from 91.9\% to 88.0\%), confirming that each model contributes to the ensemble's performance.

\textbf{Error taxonomy.}
We analyse 4,825 reasoning traces from five models across all 965 training questions.
Of these, 813 traces contain an incorrect final answer.
We check whether each model's reasoning text engages with negation patterns, exception-clause markers, and quantifier expressions present in the article, using deterministic keyword matching.
DeepSeek-R1 exhibits the highest rate of external knowledge errors (19.3\% of its incorrect traces), meaning its reasoning frequently references Civil Code articles not present in the given input.
Qwen3-235B has the highest negation detection failure rate (56.1\%), meaning it frequently fails to mention negation constructions that appear in the article.
Llama-3.3-70B is the only model with a false-positive bias (54.2\% of its errors are false positives), while the other four models tend toward false negatives.
Qwen-2.5-72B has the highest negation interpretation failure rate (71.1\%), detecting negation patterns but misinterpreting their scope or application.
These complementary error profiles explain why cross-architecture majority voting works: when Qwen3-235B fails to detect a negation, DeepSeek-R1 (with only 25.5\% detection failures) is likely to detect it and provide the correct answer, and when DeepSeek-R1 introduces external knowledge, the other four models outvote it.
Quantifier errors are corrected 73\% of the time, while negation detection failures are corrected only 40\%, as detection failures tend to occur across multiple models from the same family.
81\% of the remaining ensemble errors occur on questions where both different families and different prompts disagree, identifying the hardest questions.

\textbf{Fine-tuning comparison.}
We test whether supervised fine-tuning can match or exceed the zero-shot ensemble.
Five fine-tuning configurations are evaluated on the 258-question validation benchmark, using up to 10,525 labelled examples (965 from Task~4 and 9,560 from Task~3).
None reaches the best zero-shot expert's accuracy of 84.9\%.
The best fine-tuned result is ELYZA-8B with LoRA at 77.3\%.
The worst is Qwen-2.5-32B with QLoRA at 52.8\%, barely above random, suggesting that larger models are more susceptible to overfitting when fine-tuned on limited legal data.
This finding confirms that for statute entailment at this data scale, the broad legal reasoning capabilities encoded during pretraining are more valuable than task-specific adaptation.

\section{Pilot Task: Legal Judgment Prediction}
\label{sec:pilot}

Legal judgment prediction is a multi-level problem, since case-level outcomes depend on interactions among multiple claims, counterclaims and factual findings rather than on a single local decision. The Pilot Task requires predicting court verdicts for Japanese tort cases (TP, evaluated by accuracy) and per-claim acceptance labels (RE, evaluated by macro-averaged F1).
The training set contains 6,508 cases with 25,231 plaintiff claims and 22,458 defendant claims.
The defendant wins in 60.5\% of training cases, the median claim count per case is 4 (mean 7.3, maximum 197), and 20.3\% of cases have no defendant claims at all.
The test set contains 803 cases.

\subsection{Method}

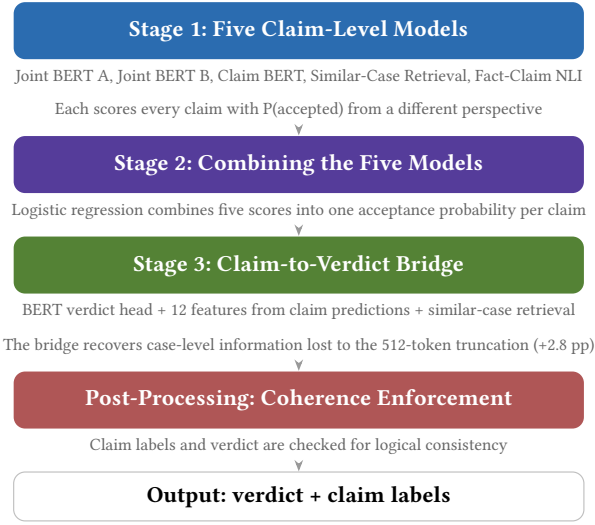
\begin{figure}[t]
\centering
\resizebox{0.95\columnwidth}{!}{%
\begin{tikzpicture}[
    node distance=0.4cm,
    >={Stealth[length=4pt]},
    stage/.style={draw=none, rounded corners=4pt, minimum height=0.7cm, text centered, fill=#1, text=white, font=\small\bfseries, minimum width=7cm},
    detail/.style={font=\scriptsize, text=black!60, align=center},
    data/.style={draw=gray!50, rounded corners=4pt, minimum height=0.6cm, text centered, fill=white, font=\small, minimum width=7cm},
    arr/.style={->, thick, color=black!40},
]
\node[stage=retrieval] (models) {Stage 1: Five Claim-Level Models};
\node[detail, below=0cm of models] (d1) {Joint BERT A, Joint BERT B, Claim BERT, Similar-Case Retrieval, Fact-Claim NLI};
\node[detail, below=0cm of d1] (d1b) {Each scores every claim with P(accepted) from a different perspective};
\node[stage=ensemble, below=0.1cm of d1b] (combo) {Stage 2: Combining the Five Models};
\node[detail, below=0cm of combo] (d2) {Logistic regression combines five scores into one acceptance probability per claim};
\draw[arr] (d1b) -- (combo);
\node[stage=verdict, below=0.1cm of d2] (bridge) {Stage 3: Claim-to-Verdict Bridge};
\node[detail, below=0cm of bridge] (d3) {BERT verdict head + 12 features from claim predictions + similar-case retrieval};
\node[detail, below=0cm of d3] (d3b) {The bridge recovers case-level information lost to the 512-token truncation (+2.8 pp)};
\draw[arr] (d2) -- (bridge);
\node[stage=llm!80, below=0.1cm of d3b] (post) {Post-Processing: Coherence Enforcement};
\node[detail, below=0cm of post] (d4) {Claim labels and verdict are checked for logical consistency};
\draw[arr] (d3b) -- (post);
\node[data, below=0.1cm of d4] (out) {\textbf{Output: verdict + claim labels}};
\draw[arr] (d4) -- (out);
\end{tikzpicture}%
}
\caption{Pilot Task: five models score each claim from different perspectives (Stage 1). A logistic regression combines their scores (Stage 2). The verdict is refined using 12 claim-derived features (Stage 3), then checked for consistency.}
\label{fig:pilot}
\end{figure}

The system has three stages (Figure~\ref{fig:pilot}).

\textbf{Stage 1: Five claim-level models.}
Each model scores every claim with a probability between 0 and 1 indicating how likely the court is to accept it.
The five models are designed to capture different aspects of claim acceptance.

\textit{Models 1 and 2: Jointly trained BERT classifiers.}
Each starts from a pretrained Japanese BERT encoder~\cite{devlin2019bert} (cl-tohoku/bert-base-japanese-v3~\cite{tohoku-bert}, 111M parameters) and adds two linear output heads: one for claim acceptance and one for the case verdict.
The claim head receives the undisputed facts paired with an individual claim.
The verdict head receives the full case text truncated to 512 tokens.
Both heads are trained simultaneously with loss $\mathcal{L} = \mathcal{L}_{\text{claim}} + 0.5 \times \mathcal{L}_{\text{verdict}}$, where both terms are binary cross-entropy.
Two instances are trained with different random seeds (42 and 7) to produce diverse predictions through convergence to different local optima.
Training uses AdamW with learning rate $2{\times}10^{-5}$ for the encoder and $2{\times}10^{-4}$ for the heads, 10\% warmup, gradient clipping at 1.0, batch size 8, up to 10 epochs with early stopping at patience 3.
Out-of-fold RE F1 scores are in the range 0.689--0.693.

\textit{Model 3: Claim-only BERT.}
This model uses the same architecture and input format as the claim head above but is trained exclusively on claim acceptance, without the verdict task.
The three supervised models achieve similar claim-level F1 (0.689--0.693), suggesting that joint training benefits verdict prediction more than claim prediction through the shared encoder.

\textit{Model 4: Dense retrieval.}
Rather than learning to classify claims directly, this model finds the most similar claims in the training data and checks whether they were accepted.
Qwen3-Embedding-0.6B (600M parameters) encodes each claim together with its case context into a dense vector.
The acceptance score is the similarity-weighted average of the gold labels of the $K$=3 nearest training claims, combined equally with a TF-IDF character n-gram baseline ($\alpha$=0.5).

\textit{Model 5: Textual entailment.}
A publicly available Japanese entailment model (akiFQC/bert-base-japanese-v3\_nli, pre-trained on three NLI datasets by its authors) scores each claim against the case facts.
The acceptance score is $\sigma(\text{entailment} - \text{contradiction})$, directly contrasting evidence for and against the claim.
This model is the weakest alone (RE F1: 0.503), since legal claim acceptance involves factors beyond textual entailment such as credibility and proportionality.
Its value lies in capturing a logical dimension that the other four models do not address: whether the facts support the claim as a matter of inference rather than pattern matching.

Alternative encoders were tested but underperformed cl-tohoku BERT\@.
DeBERTa-v2-base-japanese reduced RE F1 from 0.689 to 0.648.
ModernBERT-base (4,096 tokens), despite processing longer inputs, reduced TP accuracy from 0.708 to 0.690.

\textbf{Stage 2: Combining the five models.}
A logistic regression combines the five scores into a single acceptance probability per claim, with weights learned from 5-fold cross-validation grouped by case.
Table~\ref{tab:pilot_weights} shows the learned weights, which reflect how much we trust each model based on the validation data: the retrieval model receives the highest weight (1.856), meaning it has the most influence on the final prediction, while the NLI model receives the lowest (0.105) but non-zero, confirming a genuine contribution.

\begin{table}[t]
\centering
\caption{Logistic regression weights for combining claim-level models.}
\label{tab:pilot_weights}
\footnotesize
\begin{tabular}{lr}
\toprule
Model & Weight \\
\midrule
(4) Similar-case retrieval & 1.856 \\
(2) Joint BERT (seed B) & 1.017 \\
(1) Joint BERT (seed A) & 0.831 \\
(3) Claim-only BERT & 0.618 \\
(5) Textual entailment & 0.105 \\
\bottomrule
\end{tabular}
\end{table}

\textbf{Stage 3: Claim-to-verdict bridge.}
The BERT verdict head processes only 512 tokens, but 41\% of training cases exceed this limit.
To recover information lost to truncation, we construct twelve numerical features from the Stage 2 claim predictions (Table~\ref{tab:bridge_features}).
These features capture the balance of evidence between the two sides at different levels of granularity.
The mean acceptance difference (\texttt{re\_mean\_diff}) measures which side's claims are more likely to be accepted on average.
The top-$k$ differences (\texttt{re\_top1\_diff} through \texttt{re\_top3\_diff}) compare the strongest claims from each side, capturing whether the case has a single dominant argument or multiple supporting claims.
The threshold-based counts (\texttt{re\_count\_03}, \texttt{re\_count\_05}) measure how many claims from each side pass a moderate or strong acceptance bar, distinguishing cases with many weak claims from cases with a few strong ones.
The BERT verdict probability itself appears as a feature (\texttt{tp\_prob}), allowing the classifier to combine the encoder's direct estimate with the claim-level evidence.

A gradient-boosted decision tree takes these twelve features and produces a refined verdict estimate.
This bridge is the most impactful individual component in the entire pipeline, improving TP accuracy by 2.8 percentage points over the BERT verdict head alone.
Coefficient analysis confirms that the BERT verdict score and the mean acceptance difference between plaintiff and defendant claims contribute most to the refined prediction.
A TF-IDF retrieval component retrieves the three most similar training cases and blends their verdicts with the classifier's estimate ($\alpha$=0.3).
Since claim labels and the verdict are predicted independently, they can contradict each other, for example predicting a plaintiff win while accepting most defendant defences. Post-processing coherence rules resolve such contradictions: when the predicted verdict favours the plaintiff, no defendant defence may be newly accepted, and no more than six claim labels per party may change in a single case.

\begin{table}[t]
\centering
\caption{The twelve bridge features used in the claim-to-verdict classifier (Stage 3).}
\label{tab:bridge_features}
\setlength{\tabcolsep}{3pt}
\footnotesize
\begin{tabular}{ll}
\toprule
Feature & Description \\
\midrule
\texttt{tp\_prob} & BERT verdict score \\
\texttt{re\_mean\_diff} & Mean P acceptance prob.\ minus mean D \\
\texttt{re\_margin\_diff} & Confidence-weighted balance \\
\texttt{re\_top1\_diff} & Strongest P claim minus strongest D \\
\texttt{re\_top2\_diff} & Sum of top-2 P minus top-2 D \\
\texttt{re\_top3\_diff} & Sum of top-3 P minus top-3 D \\
\texttt{re\_count\_03} & \#P claims $\geq$0.3 minus \#D claims $\geq$0.3 \\
\texttt{re\_count\_05} & \#P claims $\geq$0.5 minus \#D claims $\geq$0.5 \\
\texttt{claim\_count} & Number of P claims minus D claims \\
\texttt{total\_claims} & Total number of claims (both sides) \\
\texttt{abs\_margin} & $|$margin\_diff$|$ (case contentiousness) \\
\texttt{abs\_tp\_anchor} & $|$tp\_prob $-$ 0.69$|$ (distance from prior) \\
\bottomrule
\end{tabular}\vspace{-0.6cm}
\end{table}

\subsection{Results and Analysis}

Table~\ref{tab:pilot} reports leaderboard entries and our component ablation.
The full system achieves 73.1\% TP accuracy (587 of 803 cases) and 68.2\% RE F1, scoring above all official entries on TP and matching the highest on RE.

Each stage provides a measurable improvement.
A single BERT model establishes the baseline at 70.9\% TP and 64.9\% RE F1.
Five-model stacking raises RE F1 to 66.9\%, a gain of 2.0 pp that confirms the models capture complementary signals about claim acceptance.
The claim-to-verdict bridge raises TP to 73.3\% (+2.8 pp over the five-model configuration without the bridge), and case-level retrieval adjusts the final system to 73.1\% TP and 68.2\% RE F1.
Among the twelve bridge features, \texttt{re\_margin\_diff} and \texttt{re\_top1\_diff} have the highest gradient-boosted feature importance.

\begin{table}[b]
\centering
\caption{Pilot Task: top leaderboard entries (official and unofficial) and component ablation.}
\label{tab:pilot}
\setlength{\tabcolsep}{3pt}
\footnotesize
\begin{tabular}{lcc}
\toprule
Configuration & TP Acc. & RE F1 \\
\midrule
\multicolumn{3}{l}{\emph{Leaderboard (official entries plus our unofficial run)}} \\
\quad \textbf{DU1 (ours, unofficial)} & \hlnum{73.1\%} & \hlnum{68.2\%} \\
\quad JNLP & 72.7\% & 64.5\% \\
\quad TortNLP4 (UIT) & 72.6\% & 68.2\% \\
\midrule
\multicolumn{3}{l}{\emph{Component ablation}} \\
\quad Single BERT model & 70.9\% & 64.9\% \\
\quad 5-model RE, no TP bridge & 70.5\% & 66.9\% \\
\quad 5-model RE + bridge, no retrieval & 73.3\% & 66.9\% \\
\quad Full system & 73.1\% & 68.2\% \\
\bottomrule
\end{tabular}
\end{table}

The retrieval model receives the highest logistic regression weight (1.856), suggesting that outcomes of similar past claims are more predictive than features from fine-tuned classifiers.
The NLI model receives the lowest weight (0.105) but is retained, confirming it contributes a unique logical entailment dimension.
The claim-to-verdict bridge recovers information lost to the BERT verdict head's 512-token truncation (41\% of cases are longer), and its +2.8 pp improvement confirms that claim-level predictions carry verdict-relevant information the encoder cannot capture alone.
Our entry was inadvertently submitted in development mode rather than leaderboard mode, and the organisers subsequently included it on the results page as an unofficial submission.

\section{Discussion}
\label{sec:discussion}

The five tasks provide a compact view of how different modelling choices behave across retrieval, entailment and judgment prediction settings in legal NLP. Three cross-task patterns emerge from our five systems.

\textbf{Classical retrieval remains competitive for legal documents.}
In Task~1, the LightGBM ranker with 34 domain features outperforms cross-encoders by 4.42 pp because documents (median 4,573 tokens) exceed encoder context windows.
In Task~3, BM25 with character bigrams retrieves the correct article as the top result for 75 of 83 gold instances without supervised training.

\textbf{Model choice matters more than pipeline complexity.}
Across Tasks~3 and~4, we observe that the choice of entailment model produces larger accuracy changes than any amount of pipeline engineering applied to a fixed model.
In Task~3, replacing Qwen2.5-72B with Qwen3-235B (using an IRAC prompt) raises accuracy from 79.3\% to 91.5\%, a gain of 12.2 pp achieved by changing nothing except the model and prompt.
By contrast, adding retrieval fusion across three systems, LLM-based reranking, and self-consistency voting to the original Qwen2.5-72B model reduces accuracy from 79.3\% to 75.6\%, a loss of 3.7 pp.
In Task~4, cross-architecture diversity contributes 4.7 pp over prompt diversity within a single model.
The error taxonomy from Task~4 confirms this: each model family has a distinct dominant failure mode, and majority voting succeeds because no single error type can accumulate a majority across diverse voters.

\textbf{Development-set assumptions can severely limit test performance.}
In Task~2, a single-selection prompt calibrated on a development set where 80\% of cases have one correct paragraph caps theoretical F1 at 0.508 on a test set averaging 2.94 correct paragraphs.
Changing only the prompt instruction, from ``select at most one'' to ``select all that entail,'' raises F1 from 0.343 to 0.555.
The entire gap between the submitted result and the best official submission is attributable to this single instruction.
This finding highlights the importance of validating task assumptions against diverse data splits, particularly in legal NLP where the number of relevant items per query can vary substantially between development and evaluation phases.
In Task~3, a related phenomenon occurs: the gold labels are exactly balanced (41 Y, 41 N), but the submitted system predicts 62--66\% Y, producing a one-sided error pattern.
Both cases illustrate the risk of calibrating output distributions on development data that does not reflect the test distribution.

\textbf{Tasks 3 and 4 provide a controlled comparison.}
These two tasks share the same 82 queries and the same Civil Code articles, differing only in whether the system retrieves articles (Task~3) or receives them from the organisers (Task~4).
On Task~4, DU1 achieves 96.3\%.
On Task~3, the same ensemble on BM25-retrieved articles achieves 86.6\%, a 9.7 pp gap not explained by retrieval failure (78/82 queries have all gold articles) but by irrelevant articles in the context confusing weaker ensemble members.
Qwen3-235B alone achieves 91.5\% on Task~3 with five retrieved articles, confirming that a strong model with a well-designed prompt tolerates moderate levels of irrelevant context.

\textbf{Reproducibility and computational footprint.}
All systems use open-weight models accessed via published checkpoints, with no proprietary model used for any prediction. For Tasks~3 and~4, every model used was released before 15 July 2025, satisfying the competition cut-off. Inference for the Task~4 ensemble is embarrassingly parallel: nine experts each issue one independent forward pass per question on the 82-question test set, with $E_9$ adding two extra samples for self-consistency. For practitioners constrained by compute, the strongest single expert ($E_9$: Qwen3-235B with self-consistency) achieves 84.9\% on the 258-question validation set, recovering a substantial part of the ensemble's gain at a fraction of the inference cost. Tables~\ref{tab:experts}, \ref{tab:prompts} and~\ref{tab:runs} and the run descriptions in Section~\ref{sec:task4}, together with the official COLIEE data, are sufficient to reconstruct each submitted run.

\section{Conclusion}
\label{sec:conclusion}

This paper presented a unified empirical study of all five COLIEE 2026 tasks through task-specific open-weight systems for retrieval, entailment, and legal judgment prediction.
Our strongest performance is obtained on Task~4, where a cross-architecture ensemble of nine open-weight models achieves 96.3\% accuracy and first place among 33 submissions, and on the Pilot Task, where a five-model pipeline with a claim-to-verdict bridge achieves 73.1\% TP accuracy as an unofficial submission, scoring above all official entries on TP.

Post-competition analysis reveals improvements that require no change to the system architecture.
For Task~2, changing a single prompt instruction from single-selection to multi-selection raises F1 from 0.343 to 0.555 in post-competition evaluation, exceeding the best official submission.
For Task~3, replacing the entailment model with Qwen3-235B using an IRAC prompt raises accuracy from 79.3\% to 91.5\% on the same retrieved articles.
Both findings were identified only after the gold labels were released, highlighting the value of thorough post-competition analysis.

The error taxonomy constructed from 4,825 reasoning traces in Task~4 provides evidence that cross-architecture diversity succeeds because each model family exhibits a characteristic failure profile.
DeepSeek-R1 introduces external legal knowledge not present in the input, Qwen3-235B fails to detect negation patterns, and Llama-3.3-70B exhibits a false-positive bias.
These complementary profiles mean that the majority vote corrects errors that no amount of prompt engineering within a single model family can address.
This finding may generalise beyond legal entailment to other classification tasks where models from different training backgrounds are available.

\section*{Acknowledgements}

Some experiments in this work used Durham University's NCC cluster, maintained by the Department of Computer Science.

\bibliographystyle{ACM-Reference-Format}
\bibliography{references}

\end{document}